\documentclass[letterpaper, 10 pt, journal, twoside]{ieeetran}
\usepackage{amsmath,amsfonts,amssymb}
\usepackage{algorithmic}
\usepackage{algorithm}
\usepackage{array}
\usepackage{multirow}
\usepackage[caption=false,font=normalsize,labelfont=sf,textfont=sf]{subfig}
\usepackage{textcomp}
\usepackage{stfloats}
\usepackage{url}
\usepackage{verbatim}
\usepackage{hyperref}
\usepackage{graphicx}
\usepackage{xcolor}
\usepackage{cite}
\newcommand{\xmark}{\ensuremath{\times}}
\newcommand{\hil}[1]{#1}

\begin{document}

\title{\textbf{Lightweight 3D Object Detection via Mamba-Based Knowledge Distillation}}

\author{
Quoc Cuong Ninh$^{1}$,
Huy Xuan Pham$^{2}$,
Anh Tung Nguyen$^{3}$, and
Dinh Hoan Trinh$^{1,*}$%

\thanks{$^{1}$Quoc Cuong Ninh and Dinh Hoan Trinh are with Viettel AI, Viettel Group, Vietnam.
{\tt\footnotesize \{cuongnq23, hoantd5\}@viettel.com.vn}}%
\thanks{$^{2}$Huy Xuan Pham is with the Artificial Intelligence in Robotics Laboratory (AiR Lab), Department of Electrical and Computer Engineering, Aarhus University, 8000 Aarhus C, Denmark, and also with Upteko ApS, Denmark. {\tt\footnotesize huy.pham@ece.au.dk}}%
\thanks{$^{3}$Anh Tung Nguyen is with VIST, Viettel Group, Vietnam. {\tt\footnotesize tungna@viettel.com.vn}}%
\thanks{$^{*}$Corresponding author.}
\thanks{\textbf{Project Page:} \url{https://lightweight-mamba-kd.github.io/home/}}

}

\maketitle

\begin{abstract}
3D object detection using light detection and ranging (LiDAR) sensors requires a balance between accuracy and computational efficiency for onboard perception in autonomous driving and robotic navigation. Many existing LiDAR-based detection methods employ complex architectures to extract features, integrating large amounts of contextual information to enhance accuracy. This often results in significant computational costs, leading to suboptimal performance on resource-constrained embedded devices. In this study, we propose a knowledge distillation framework that transfers object-level voxel representations from a strong teacher model to lightweight student models through selective voxel-space feature alignment. Taking advantage of the linear-time sequence model with selective state spaces (Mamba), we design a multi-branch Mamba teacher backbone and a box-aware feature transfer mechanism that aligns spatially corresponding voxel features between teacher and student networks through a Mamba-based projection module.
 Experimental results on both a public dataset and real-world data show that our approach significantly reduces computational load while maintaining competitive accuracy compared with state-of-the-art methods.
\end{abstract}

\section{Introduction}
LiDAR-based 3D object detection is a fundamental task for many applications in autonomous driving and intelligent robotic systems.

Unlike depth or stereo cameras, which capture photometric and dense geometric information but can be affected by background clutter and have limited fields of view (FoV), 3D LiDAR utilizes laser beams to reconstruct detailed spatial representations in a large field of view, with 3D points focused on prominent objects. This characteristic enhances the ability to capture the geometric properties of objects through sparse point clouds. In recent years, large-scale LiDAR-based 3D perception datasets, such as KITTI\cite{kitti}, NuScenes \cite{nuscenes}, and Waymo Open Dataset \cite{waymo}, have been proposed to facilitate emerging sophisticated models leveraging architectures such as transformers\cite{transformer}. These models improve the accuracy of 3D object detection, but often at the cost of substantial computational complexity, requiring powerful GPU processing and posing challenges for resource-constrained autonomous vehicles and robots in real-time applications.

\begin{figure}[t]
    \centering
    \includegraphics[width=0.98\columnwidth]{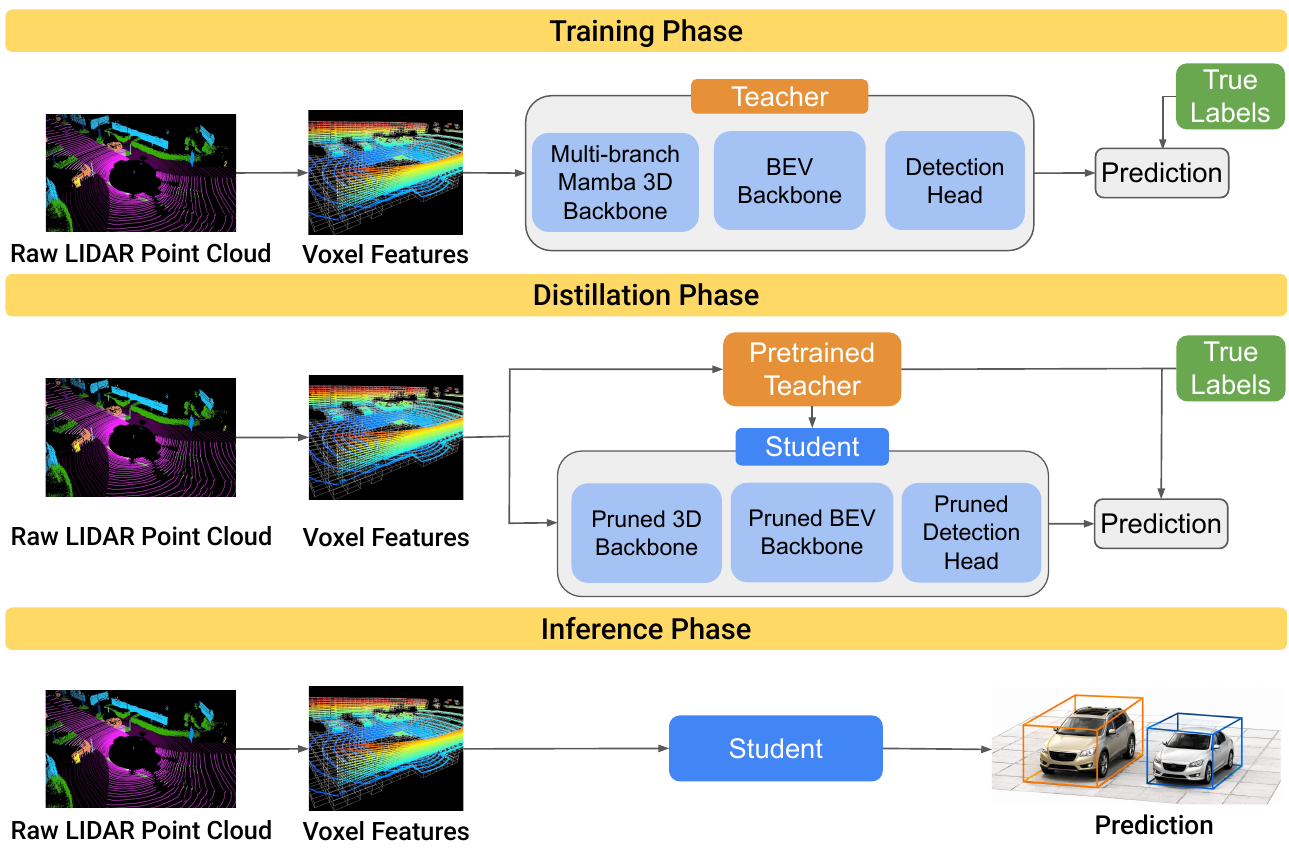}
    \caption{\hil{Overview of the proposed 3D object detection pipeline. The framework consists of a multi-branch Mamba-based teacher model and a compact student model trained via knowledge distillation, including training, distillation, and inference phases.}}
    \label{fig:pipeline}
\end{figure}

To address computational complexity, model compression techniques have gained attention. Knowledge distillation (KD), which trains a lightweight student network under the guidance of a more complex teacher network, has proven effective in various vision tasks, including both 2D and 3D detection. KD enables the student to inherit classification and deep feature knowledge from the teacher, thus improving accuracy compared to standalone training. However, applying KD to 3D point cloud detection presents unique challenges, for instance, handling sparse data distributions and multidimensional outputs, such as bounding box coordinates and orientations.

In the point cloud-based detection problem, capturing the relationships between 3D points is crucial for accurate object recognition. Recent approaches have treated spatial points as sequential data for feature aggregation. A line of work \cite{liu2024lion, voxelmamba} utilizes state-space models with a selective mechanism from Mamba \cite{s6}, offering superior spatiotemporal feature extraction from sparse LiDAR data. Unlike other sequence models such as transformer architectures, which incur quadratic complexity, Mamba’s linear complexity makes it attractive for deployment on resource-constrained robotic platforms, where computational efficiency is a critical requirement.

This study focuses on improving feature transfer between teacher and student models for LiDAR-based 3D object detection. In particular, we investigate how sparse voxel representations produced by Mamba-based detectors can be more effectively distilled through spatially aligned feature matching and feature-space projection. To achieve a balance between computational efficiency and contextual awareness in 3D object detection from LiDAR data, we propose a novel architecture that incorporates multiple branches to enhance global context capture during feature extraction. To address the challenge in model complexity and latency reduction, a knowledge distillation strategy is proposed, in which lightweight student models are developed through selective pruning of the original architecture. The cost-performance ratio (CPR) metric \cite{spartialkd} is adopted to quantify the trade-off between computational cost and model quality of the student models to find the optimal architectures.  Furthermore, a box-aware distillation mechanism is introduced to transfer selective voxel features within object regions from teacher to student models through spatially corresponding voxel alignment and feature-space projection, improving detection accuracy without increasing inference cost.

\section{Related work}
\label{sec:Related}
\subsection{LiDAR-based 3D Object Detection}
The LiDAR-based 3D object detection task aims to localize and classify objects in 3D space by predicting bounding boxes from point cloud data. Early methods transformed point clouds into structured representations, such as voxel grids, to leverage convolutional neural networks (CNNs). 
For instance, SECOND \cite{yan2018second} utilized sparse 3D convolutions to process voxelized point clouds, achieving robust performance. PointPillars \cite{lang2019pointpillars} further reduced computational complexity by collapsing the height dimension into "pillars," thereby enabling faster inference suitable for real-time applications. Point-based methods, such as PointNet++ \cite{qi2017pointnetplusplus}, directly process raw point clouds, exploiting order invariance and local point relationships; however, they often suffer from high computational costs for large-scale scenes. Hybrid approaches like PV-RCNN \cite{pvrcnn} combine voxel-based feature extraction with point-based refinement, while CenterPoint \cite{yin2021centerbased} employs an anchor-free head for efficient object center localization, which is widely adopted in autonomous driving. More recently, sequence-aware methods like DSVT \cite{wang2023dsvt} leverage transformer architectures\cite{transformer} to model spatial-temporal relationships, achieving high accuracy but incurring significant computational overhead due to quadratic complexity. These methods highlight the trade-off between accuracy and computational efficiency, motivating our work to develop a lightweight model that maintains high accuracy for real-time applications.

\subsection{Mamba-based Architectures in Vision and 3D Detection}
\begin{figure*}[t]
    \centering
    \includegraphics[width=2\columnwidth]{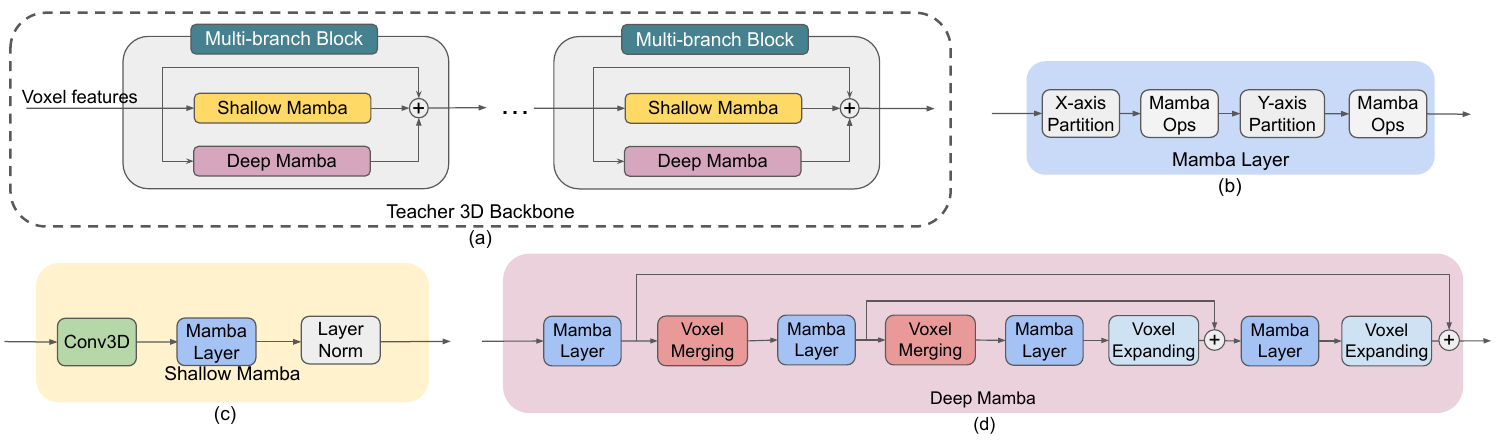}

    \caption{Architecture of the multi-branch Mamba 3D backbone used as the teacher model. (a) Overall structure with stacked multi-branch blocks (MBBs). (b) Mamba layer design. (c) Shallow Mamba branch. (d) Deep Mamba branch.}
    \label{fig:teacher}
\end{figure*}

In recent years, state-space models (SSM) have garnered significant attention for their efficacy in sequence modeling. Several SSM-based architectures have been proposed \cite{LSSL,DSS,S4D} to handle sequential data across various tasks. The recently introduced Mamba architecture \cite{s6}, which incorporates time-varying parameter selection into SSM, enhances training and inference efficiency, offering a compelling alternative to Transformers for long-sequence data. Mamba-based models have shown remarkable progress in 2D image processing, with architectures such as Vim \cite{vim} and VMamba \cite{vmamba} demonstrating robust performance. This trend has extended to 3D spatial processing, where Mamba exhibits promising results in handling discrete spatial point sequences. For instance, LION \cite{liu2024lion} employs Mamba-based recurrent processing on voxelized point clouds, achieving competitive accuracy compared to Transformer-based detectors such as DSVT \cite{wang2023dsvt}. 
Similarly, Voxel Mamba \cite{voxelmamba} leverages Mamba’s ability to compress sequential signals into latent vectors, improving spatiotemporal feature extraction from LiDAR data. However, Mamba-based models may struggle to capture global context as effectively as transformers due to their linear sequence processing nature.

\subsection{Knowledge Distillation in 3D Object Detection}\label{SCM}

To transfer knowledge from a complex teacher model to a lightweight student model, KD methods have recently been used to reduce inference costs while preserving accuracy. In 3D object detection, KD must address the complexity of spatial data and diverse outputs, such as 3D bounding boxes and orientations. 
Notable approaches include itKD \cite{itkd}, which proposes an autoencoder‑style knowledge distillation framework with channel‑wise compression/decompression interchanges and a head relation‑aware self‑attention loss to transfer rich 3D point‑cloud detection features from teacher to student, yielding efficient yet accurate lightweight models on datasets like Waymo and nuScenes. PointDistiller \cite{zhang2022pointdistillerstructuredknowledgedistillation} employs a structured feature-based knowledge distillation method for 3D point cloud object detectors, leveraging dynamic graph convolutions on important local voxels with a reweighted loss strategy to produce compact models (e.g., 4× smaller PointPillars) that often surpass their teachers in both bird’s eye view (BEV) and 3D detection accuracy. 
SparseKD \cite{spartialkd} conducts a systematic study of KD for LiDAR-based models, introducing Enhanced Logit KD and Teacher-guided Initialization, achieving significant mAP improvements on NuScenes. However, these methods primarily rely on CNN-based architectures to transfer features from complex teacher models to lightweight student models, struggle to capture long-range dependencies in sparse point clouds, incur high computational costs from operations like graph convolutions, and depend on handcrafted loss functions that may not generalize well.

\section{Methodology} \label{sec:Method}

\subsection{Overview}

We propose a framework that leverages knowledge distillation to transfer knowledge from a high-capacity, multi-branch Mamba-based teacher model to lightweight student models, as illustrated in Fig. \ref{fig:pipeline}. Both the teacher and student models consist of a 3D backbone (3DB), a BEV backbone (BB), and a detection head (DH), maintaining a consistent pipeline for voxel-based 3D object detection. For the student models, we focus on pruning the teacher architecture to achieve efficient, lightweight designs. Additionally, we describe the knowledge distillation process, which transfers latent feature representations from the output of the teacher model’s Mamba blocks to the student model’s Mamba blocks, 

facilitating effective feature transfer between teacher and student models in Mamba-based 3D detectors.

\subsection{Teacher Architecture}

A novel Multi-branch Mamba 3D Backbone is proposed to enhance the representational capacity for LiDAR-based 3D perception (Fig.~\ref{fig:teacher}-a). The core component of the teacher architecture is a Mamba layer (Fig.~\ref{fig:teacher}-b). The Mamba layer models long-range dependencies among voxel features through structured sequential processing. Given a sparse voxel feature matrix, voxel features are first reorganized into ordered sequences using a 3D sparse window partition strategy \cite{liu2024lion}. Specifically, voxels are grouped and arranged along one spatial direction (X-axis partition), forming a structured sequence that preserves spatial locality within each group. This ordered sequence is then processed by a Mamba operator \cite{s6}, which performs linear recurrent state updates to efficiently aggregate contextual information. To capture complementary spatial dependencies, the partitioning and Mamba interaction are repeated along an orthogonal direction (Y-axis partition). After sequential processing along both directions, the updated features are reshaped back to the original voxel layout.

The backbone is composed of multiple multi-branch blocks (MBBs), each integrating two parallel branches: a Shallow Mamba branch and a Deep Mamba branch. The Shallow Mamba branch (Fig.~\ref{fig:teacher}-c) consists of a sequence of standard Mamba layers combined with Layer Normalization and a 3D convolutional layer to extract spatial 3D features. It focuses on learning low-level features, prioritizing computational efficiency. The Deep Mamba branch (Fig.~\ref{fig:teacher}-d) employs a more complex structure, incorporating voxel merging and voxel expanding techniques from LION \cite{liu2024lion}, interleaved with Mamba layers. This process enables the model to learn higher-level features by reducing and expanding the voxel space, thereby capturing intricate spatial relationships. The voxel merging and voxel expanding operations are adopted directly from LION without architectural modification. The outputs from the shallow and deep branches are concatenated and fused through a residual connection before being processed by the BEV backbone. This backbone compresses these features to generate a 2D spatial representation, which is then fed into the detection head for final predictions. By jointly leveraging these complementary branches within each block, the architecture strengthens multi-scale feature representation from sparse 3D point clouds.

\subsection{Student Architecture}

Lightweight student models are developed by reducing the computational complexity of the teacher architecture. For each module of the teacher network (3DB, BB, DH), $width$ refers to the number of feature channels, while $depth$ denotes the number of Mamba layers or convolutional layers. The teacher model serves as the baseline configuration with $width = 1$ and $depth = 1$, and the student models are generated by scaling these dimensions with different ratios, as summarized in Table~\ref{tab:pruning}. To isolate the effect of compressing the Mamba-based 3D backbone, the depth of the BEV backbone is kept fixed across all student variants. Only the channel width of the BEV backbone is adjusted when necessary to maintain compatibility with the compressed 3D backbone.

To select the most efficient student models, we apply a pruning strategy guided by the CPR concept \cite{spartialkd}. The original CPR metric combines the ratio of activations and mAPH to evaluate the trade-off between model efficiency and detection performance. In this work, we introduce two modifications. First, we replace activations with measured inference latency, which more directly reflects deployment efficiency on embedded robotic platforms. Second, we use mAP as the detection-quality metric to maintain consistency with the evaluation metrics adopted throughout this paper. The revised CPR is defined as follows:

\begin{equation}
   CPR = 0.5 \times \left(1 - \frac{\text{latency}_s}{\text{latency}_t}\right) + 0.5 \times (\frac{\text{mAP}_s}{\text{mAP}_t})^3,
\end{equation}

To compute CPR, we estimate model complexity metrics, including the number of parameters, activations, and inference latency measured on a single NVIDIA A4000 GPU. The CPR is computed for each student model to assess pruning efficiency, where a higher CPR indicates a more favorable trade-off between accuracy and computational cost.

\subsection{Mamba Adaptive Distillation}

A novel knowledge distillation framework is proposed for Mamba-based 3D object detection, combining head output distillation with a specialized latent space KD loss targeting Mamba block outputs. This approach enhances knowledge transfer from a high-capacity teacher model to lightweight student models, utilizing Mamba’s sequential processing capabilities to capture contextual information in voxelized LiDAR point clouds. Drawing on soft target distillation \cite{Hinton2015DistillingTK}, we apply KD to the detection head outputs of the teacher and a student model, which comprise classification, regression, and heatmap components. The head KD loss is formulated as:

\begin{equation} \mathcal{L}_{\text{KD}}^{\text{head}} = \mathcal{L}_{\text{KD}}^{\text{cls}} + \mathcal{L}_{\text{KD}}^{\text{reg}} + \mathcal{L}_{\text{KD}}^{\text{hm}}, \end{equation}

where:

\begin{equation}
\mathcal{L}_{\text{KD}}^{k} = \mathbb{E} \left\| \mathbf{p}_{k}^{t} \right\|_2, 
\quad 
k \in \{\text{cls}, \text{reg}, \text{hm}\}.
\end{equation}

Here, $(\mathbf{p}_{\text{cls}}), (\mathbf{p}_{\text{reg}})$, and $(\mathbf{p}_{\text{hm}})$ represent the classification, bounding box regression, and heatmap outputs, respectively, with superscripts (s) and (t) denoting the student and teacher models. Outputs are aligned by matching bounding boxes with the highest Intersection over Union (IoU).

\begin{figure}[b]
    \centering
    \includegraphics[width=0.9\columnwidth]{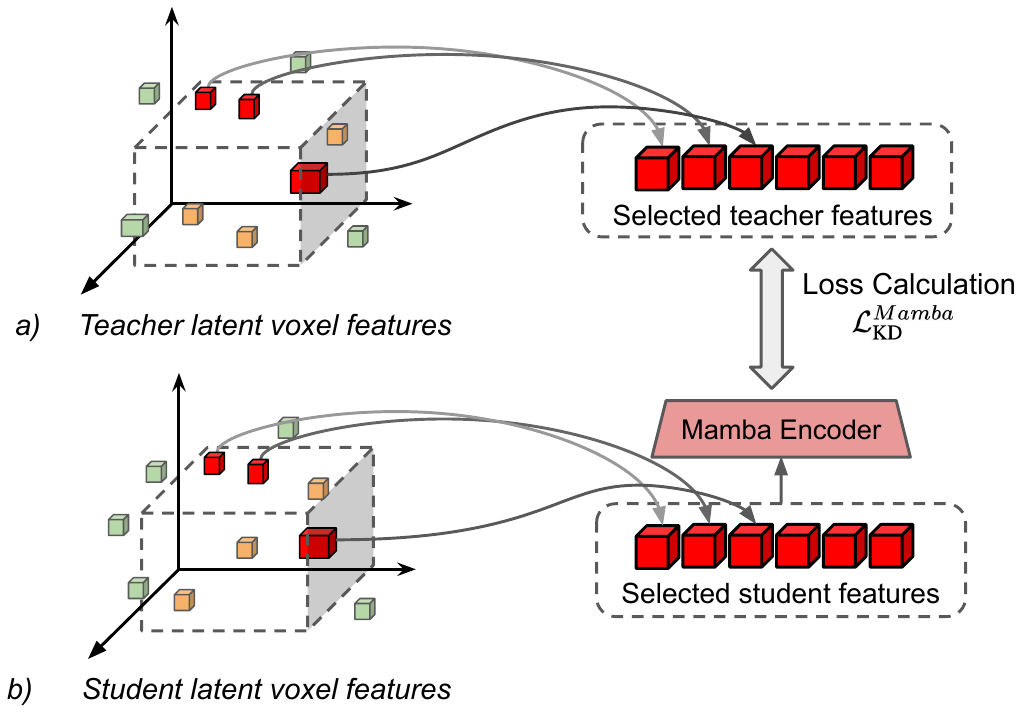}
    \caption{Box-aware latent feature distillation in 3D voxel space.
    The dashed 3D bounding box denotes the ground-truth box of object $i$. For each object, voxel features within the box are spatially filtered from both teacher and student latent voxel feature maps (red and orange cubes). Spatially corresponding voxels (red cubes) are matched in a coordinate-consistent manner. The selected student features are further projected through a lightweight Mamba encoder to match the teacher dimensionality. The aligned object-aware representations $\mathcal{F}_T[i, :]$ and $\mathcal{F}_S[i, :]$ are then optimized using a distillation loss.
    }
    \label{fig:mamba_feat}
\end{figure}

Conventional feature map distillation, as proposed by \cite{romero2015fitnetshintsdeepnets}, transfers knowledge from intermediate layers using a loss such as:

\begin{equation} \mathcal{L}_{\text{KD}}^{\text{feat}} = \left| F_T - F_S \right|_2, \end{equation}

where $(F_T)$ and $(F_S)$ are feature maps from the teacher model and a student model, respectively. While feature distillation can be applied to a variety of 3D detection architectures, directly transferring sparse voxel representations between teacher and student networks remains challenging due to differences in feature dimensionality, sparse spatial layouts, and backbone architectures. To address these challenges, we introduce a novel latent space KD loss that operates on the output of Mamba blocks, where sequential state representations converge. Unlike prior methods focusing on BEV features in 3D object detectors, our approach targets features extracted after Mamba-based voxel feature encoding, leveraging Mamba’s ability to capture long-range contextual dependencies in sparse 3D point clouds. To enhance efficiency, a selective feature transfer mechanism is proposed to distill only the features corresponding to voxel indices within ground-truth 3D bounding boxes, prioritizing semantically relevant object-related features and mitigating background noise, as demonstrated in Fig. \ref{fig:mamba_feat}. Teacher and student features are aligned by matching shared voxel indices for architectural consistency.

Formally, let $\mathcal{F}_T \in \mathbb{R}^{N \times C_T}$ and $\mathcal{F}_S \in \mathbb{R}^{N \times C_S}$ denote the feature tensors output by the Mamba blocks of the teacher and student model, respectively, where $N$ is the total number of voxels and $C_T, C_S$ are the feature dimensions of the teacher and student model. Let $\mathcal{B} = \{b_1, b_2, \ldots, b_M\}$ represent the set of $M$ ground-truth 3D bounding boxes, and $\mathcal{I}_{b_i} \subseteq \{1, 2, \ldots, N\}$ denote the set of voxel indices within the $i$-th bounding box $b_i$. We define the common voxel indices across both models as $\mathcal{I}_{\text{common}} = \mathcal{I}_T \cap \mathcal{I}_S$, where $\mathcal{I}_T$ and $\mathcal{I}_S$ are the sets of valid voxel indices in the teacher and student feature tensors, respectively, within the union of all bounding boxes $\bigcup_{i=1}^M \mathcal{I}_{b_i}$. The knowledge distillation loss is then computed only over the features corresponding to these common indices:
\begin{equation}
\mathcal{L}_{\text{KD}}^{Mamba} = \frac{1}{|\mathcal{I}_{\text{common}}|} \sum_{i \in \mathcal{I}_{\text{common}}} \left\| \mathcal{F}_T[i, :] - \phi(\mathcal{F}_S[i, :]) \right\|_2^2,
\end{equation}

where $\mathcal{F}_T[i,:]$ and $\mathcal{F}_S[i,:]$ denote the teacher and student feature vectors at voxel index $i$, respectively. The function $\phi(\cdot)$ represents a feature projection module that maps the student feature representation to the teacher feature space. In the proposed method, $\phi(\cdot)$ is implemented using a lightweight Mamba encoder, although alternative projection modules are also investigated in the ablation study. The operator $\|\cdot\|_2^2$ denotes the squared L2 distance. This loss ensures that the student model learns to mimic the teacher's feature representations for voxels within the ground-truth bounding boxes, enhancing the transfer of object-specific contextual knowledge.

\begin{figure}[b]
    \centering
    \includegraphics[width=1\columnwidth]{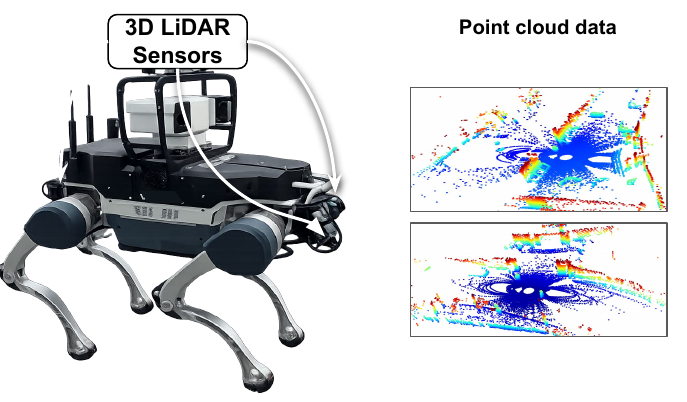}
    \caption{Legged robot equipped with 3D LiDAR sensors and representative point cloud samples from the collected Livox-Legged dataset.
    }
    \label{fig:robot}
\end{figure}

By focusing on Mamba-extracted voxel features and selectively transferring knowledge based on ground-truth bounding box indices, our method addresses the challenges of sparse point cloud data and ensures robust generalization. This approach also reduces computational overhead by excluding irrelevant background features and aligns the feature spaces of the teacher and the student model, leading to improved accuracy and efficiency in 3D object detection tasks.

We define a composite objective loss that integrates knowledge distillation from the teacher’s Mamba block output and detection head. The final objective loss is formulated as:

\begin{equation} \mathcal{L} = \mathcal{L}_{\text{gt}} + \alpha \cdot \mathcal{L}_{\text{KD}}^{\text{Mamba}} + \beta \cdot \mathcal{L}_{\text{KD}}^{\text{head}},
\end{equation}

where weights $(\alpha)$ and $(\beta)$ balance the contributions of the KD losses, typically set to 1.0 and 0.5, respectively, to prioritize ground truth supervision while ensuring effective knowledge transfer. This composite loss leverages Mamba’s sequential modeling and head output alignment to enhance the student’s performance on sparse 3D point clouds.

\section{Experiments} \label{sec:Experiment}

\subsection{Dataset and Metrics}

The proposed 3D object detection framework is evaluated on both a public benchmark and a real-world proprietary dataset. First, we conduct experiments on the LiDAR-only track of the nuScenes dataset~\cite{nuscenes}, a large-scale autonomous driving benchmark comprising 1,000 scenes and approximately 28,000 annotated LiDAR keyframes across 10 object categories. We follow the official split, using 700 scenes for training and 150 scenes for validation. In addition, we collect a real-world proprietary dataset, termed Livox-Legged, using multiple Livox Mid-360 LiDAR sensors mounted on a legged robot (Fig. \ref{fig:robot}). The point clouds from all sensors are spatially calibrated and merged into a unified representation. The dataset contains 2,000 training scenes and 500 testing scenes, covering diverse outdoor navigation scenarios. It includes three object categories: pedestrians, cars, and motorcyclists. Across the training and testing splits, it contains 13,933 pedestrian instances, 3,971 car instances, and 1,617 motorcyclist instances. Each object is annotated with a 3D bounding box in the calibrated LiDAR coordinate system.

For evaluation on the nuScenes validation set, we adopt the standard metrics, including mean average precision (mAP) and NuScenes Detection Score (NDS). We further report the five true-positive error metrics: average translation error (ATE), average scale error (ASE), average orientation error (AOE), average velocity error (AVE), and average attribute error (AAE), which measure localization, scale, orientation, velocity, and attribute prediction accuracy, respectively. For a fair comparison with state-of-the-art methods, latency is measured on an NVIDIA RTX A4000 GPU under identical inference settings. On the Livox-Legged dataset, we report mAP and inference latency to evaluate both detection accuracy and deployment efficiency. {Following the nuScenes evaluation protocol, true-positive metrics are computed using center-distance matching with a threshold of $d=2$ m for all classes.}
Latency is measured on an NVIDIA Jetson Orin NX platform to reflect real-world onboard robotic deployment.

\subsection{Experimental Settings}
\begin{table*}[t]
    \centering
    \caption{Model Compression Results with Varying Pruning Ratios and Corresponding CPR Metrics. Note that 3DB, BB, and DH Stands for 3D Backbone, BEV Backbone, and Detection Head}
    \resizebox{0.8\textwidth}{!}{
        \begin{tabular}{c|ccc|cc|ccc|c|c}
        \hline
        \multirow{2}{*}{Detector} & \multicolumn{3}{c|}{Width} & \multicolumn{2}{c|}{Depth} & \multicolumn{3}{c|}{Efficiency} & \multirow{2}{*}{mAP} & \multirow{2}{*}{CPR}\\
            & 3DB & BB & DH & 3DB & BB & Params (M) & Activations (M) & Latency (ms) & & \\
        \hline
        \multirow{1}{*}{Teacher}
        & 1.00 & 1.00 & 1.00 & 1.00 & 1.00 & 23.52  & 339.2 & 199 & 60.68  & -  \\
        
        \multirow{1}{*}{student-a}
        & 1.00 & 0.50 & 0.50 & 1.00 & 1.00 & 11.60 & 238.4 & 158 & 56.70 & 0.55  \\
        
        \multirow{1}{*}{student-b}
        & 0.75 & 0.50 & 0.50 & 1.00 & 1.00 & 7.40 & 189.7 & 147 & 54.76 & 0.53 \\
        
        \multirow{1}{*}{student-c}
        & 0.50 & 0.25 & 0.25 & 1.00 & 1.00 & 3.18 & 112.6 & 105 & 53.73 & 0.62  \\
        
        \multirow{1}{*}{\textbf{student-d}}
        & 1.00 & 1.00 & 1.00 & 0.50 & 1.00 & 11.04  & 315.4 & 108 & 58.54 & \textbf{0.72}  \\
    
        \multirow{1}{*}{student-e}
    
        & 0.75 & 0.50 & 0.50 & 0.50 & 1.00 & 4.44  & 172.3 & 77 & 54.50 & 0.70 \\
    
        \multirow{1}{*}{student-f}
    
        & 0.50 & 0.25 & 0.25 & 0.50 & 1.00 & 1.74  & 107.6 & 51 & 50.73 & 0.69  \\

        \multirow{1}{*}{\textbf{student-g}}
    
        & 1.00 & 1.00 & 1.00 & 0.25 & 1.00 & 10.50  & 297.7 & 97 & 57.48 & \textbf{0.73}  \\

        \multirow{1}{*}{\textbf{student-h}}
    
        & 0.75 & 0.50 & 0.50 & 0.25 & 1.00 & 4.11  & 161.0 & 71 & 54.56 & \textbf{0.72}  \\
    
        \hline
        \end{tabular}
    }
    \label{tab:pruning}
    \end{table*}

\newcommand{\indentcell}[1]{\hspace{1em}#1}

\begin{table*}
\centering
\caption{Comparison of Knowledge Distillation Methods on the NuScenes Validation Set. All Methods Are Applied to the Student-d Variant Trained with 20\% of the NuScenes Training Data}
\resizebox{0.8\textwidth}{!}{
\begin{tabular}{l|c|c|c|c|c|c|c}
\hline
\multicolumn{1}{c|}{KD schemes} & 
mATE~$\downarrow$ & 
mASE~$\downarrow$ & 
mAOE~$\downarrow$ & 
mAVE~$\downarrow$ & 
mAAE~$\downarrow$ & 
mAP~$\uparrow$ & 
NDS~$\uparrow$ \\

\hline
Teacher model w/o KD & 0.289 & 0.258 & 0.307 & 0.307 & 0.195 & 60.68 & 66.79\\
Student model w/o KD & 0.296 & 0.260 & 0.321 & 0.327 & 0.198 & 58.54 & 65.25 \\
Student model + FitNet\cite{romero2015fitnetshintsdeepnets} & 0.280 & 0.258 & 0.290 & 0.295 & 0.194 & 60.93 & 67.30\\
Student model + Mimic\cite{mimic} & 0.283 & 0.259 & 0.290 & 0.304 & 0.194 & 60.27 & 66.83\\
Student model + FG\cite{fg} & 0.281 & 0.258 & 0.287 & 0.296 & 0.193 & 61.00 & 67.36\\
Student model + SparseKD\cite{spartialkd} & 0.282 & 0.259 & \textbf{0.283} & 0.284 & 0.190 & 61.07 & 67.54\\
Student model + PointDistiller\cite{zhang2022pointdistillerstructuredknowledgedistillation} & 0.289 & 0.263 & 0.308 & 0.306 & 0.195 & 59.84 & 66.32\\
Student model + itKD\cite{itkd} & 0.281 & 0.259 & 0.295 & 0.297 & 0.193 & 60.35 & 66.92\\
\textbf{Student model + Ours} & \textbf{0.276} & \textbf{0.256} & 0.286 & \textbf{0.282} & \textbf{0.190} & \textbf{63.00} & \textbf{68.59}\\
\hline
\end{tabular}
}
\label{tab:compare_kd}
\end{table*}

All models are implemented within the OpenPCDet framework~\cite{openpcdet2020} using PyTorch as the deep learning backend. For consistency, both teacher and student models are trained for 36 epochs with a batch size of 16. The teacher 3D backbone contains four multi-branch blocks. We adopt the same BEV backbone and detection head design as DSVT~\cite{wang2023dsvt}. The point cloud range, voxelization grid size, data augmentation strategies, optimizer, and learning rate schedule follow the configuration of DSVT~\cite{wang2023dsvt}. Training is conducted on four NVIDIA L40 GPUs. All experiments are initialized with a fixed random seed to ensure reproducibility and fair comparison. For both nuScenes and Livox-Legged training, we employ the same point cloud preprocessing pipeline and training protocol.
\subsection{Results and Analysis}
\begin{table}[t]
\centering
\caption{
Comparison of LiDAR-based 3D Detectors on the NuScenes Validation Set Using the Full Training Set
}

\resizebox{0.5\textwidth}{!}{
\begin{tabular}{c|c|c|c}
\hline
Models & Params (M) $\downarrow$ & Latency (ms) $\downarrow$ & mAP (\%) $\uparrow$\\
\hline
CenterPoint$^\dagger$ & 5.98 & \textbf{26}  & 59.2 \\
VoxelNext$^\dagger$ & 19.05 & 46 & 60.5  \\
Transfusion-L$^\dagger$ & 8.38 & 67 & 64.6\\
DSVT$^\ast$ & 8.20 & 84 & 66.4\\
Voxel Mamba$^\ast$ & 22.35 & 230 & 67.5\\
LION$^\ast$ & 16.52 & 185 & 68.0\\
Teacher Model (Ours) & 23.52 & 199 & \textbf{68.2} \\
Student-d (Ours) & 11.04 & 108  & 67.9 \\
Student-g (Ours) & 10.50 & 97  & 67.1 \\
Student-h (Ours) & \textbf{4.11} & 71  & 65.6 \\
\hline
\end{tabular}
}
\vspace{2pt}
\\
{\footnotesize 
$^\dagger$ Using OpenPCDet pretrained weights. 
$^\ast$ Reported in original papers.}
\label{tab:comparesota}
\end{table}

\begin{table}[t]
\centering
\caption{
\hil{Per-class AP (\%) and Inference Latency on the Livox-Legged Test Set.
All Models are Trained from Scratch on the Livox-Legged Dataset.
Latency is Measured on an NVIDIA Jetson Orin NX with Batch Size 1}
}
\resizebox{0.48\textwidth}{!}{
\begin{tabular}{l|c|c|c|c|c}
\hline
Method & Latency (ms) & Car (\%) & Pedestrian (\%) & Motorcyclist (\%) & mAP (\%) \\
\hline
CenterPoint~\cite{yin2021centerbased} & \textbf{142} & 92.55 & 36.60 & 32.14 & 53.76 \\
DSVT~\cite{wang2023dsvt} & 425 & 91.64 & 35.84 & 44.78 & 57.42 \\
LION~\cite{liu2024lion} & 928 & 94.24 & 33.48 & 35.02 & 54.25 \\
\textbf{Student-h (Ours)} & 350 & \textbf{99.48} & \textbf{47.03} & \textbf{61.12} & \textbf{69.21} \\
\hline
\end{tabular}
}
\label{tab:realworld}
\end{table}

\textbf{Effectiveness of the proposed KD method.}
Table~{\ref{tab:pruning}} shows the model compression results with varying pruning configurations. The teacher and student models are trained using 20\% of the nuScenes training set to accelerate experimental iteration, and performance is evaluated on the nuScenes validation set using mAP. Reducing the depth of the 3DB module (i.e., the number of Mamba layers) leads to a substantial reduction in inference latency, while the total number of activations decreases only marginally. This observation suggests that, in Mamba-based architectures, latency is not strictly proportional to activation count and is strongly influenced by the computational characteristics of the Mamba operator. Based on this analysis, we select the \textit{d}, \textit{g}, and \textit{h} variants (highlighted in Table {\ref{tab:pruning}}) for subsequent knowledge distillation experiments.

Table~\ref{tab:compare_kd} compares the \textit{student-d} model trained with only 20\% of the nuScenes training data against existing KD approaches. Our Mamba-based distillation framework achieves superior performance, with the student model outperforming other KD methods as well as the non-distilled teacher baseline. This result demonstrates that the proposed KD loss effectively transfers structural and contextual knowledge while compensating for model pruning. 

\textbf{Comparison with state-of-the-art methods.}
We further compare our models with representative LiDAR-based 3D detectors, including CenterPoint~\cite{yin2021centerbased}, VoxelNext~\cite{voxelnext}, Transfusion-L~\cite{transfusionl}, DSVT~\cite{wang2023dsvt}, VoxelMamba~\cite{voxelmamba}, and LION~\cite{liu2024lion} on the nuScenes validation set. All proposed models in Table~{\ref{tab:comparesota}} are trained using the full nuScenes training set. Baseline results marked with {\(^{\dagger}\)} are obtained from publicly available OpenPCDet pretrained models, while results marked with {\(^{\ast}\)} are reported in the corresponding original papers. Therefore, these baselines may use different training schedules, implementation details, or experimental protocols. Table~{\ref{tab:comparesota}} is intended to position the proposed method relative to representative state-of-the-art detectors rather than to serve as a fully controlled benchmark reproduction.
Parameters and latency are measured under identical hardware settings for fair comparison. As shown in Table~\ref{tab:comparesota}, the proposed multi-branch Mamba teacher 
achieves competitive detection accuracy among the compared methods.
Notably, \textit{student-d} retains near-teacher performance while reducing model size and inference time by approximately half. The most compact variant, \textit{student-h}, maintains competitive accuracy while 
improving efficiency, outperforming several established baselines despite substantially fewer parameters. These results validate the effectiveness of the proposed distillation strategy in producing compact yet high-performing detectors suitable for real-world deployment.

\begin{figure*}[htbp!]
\centering
\includegraphics[width=0.9\textwidth, height=4.3cm, keepaspectratio=false]{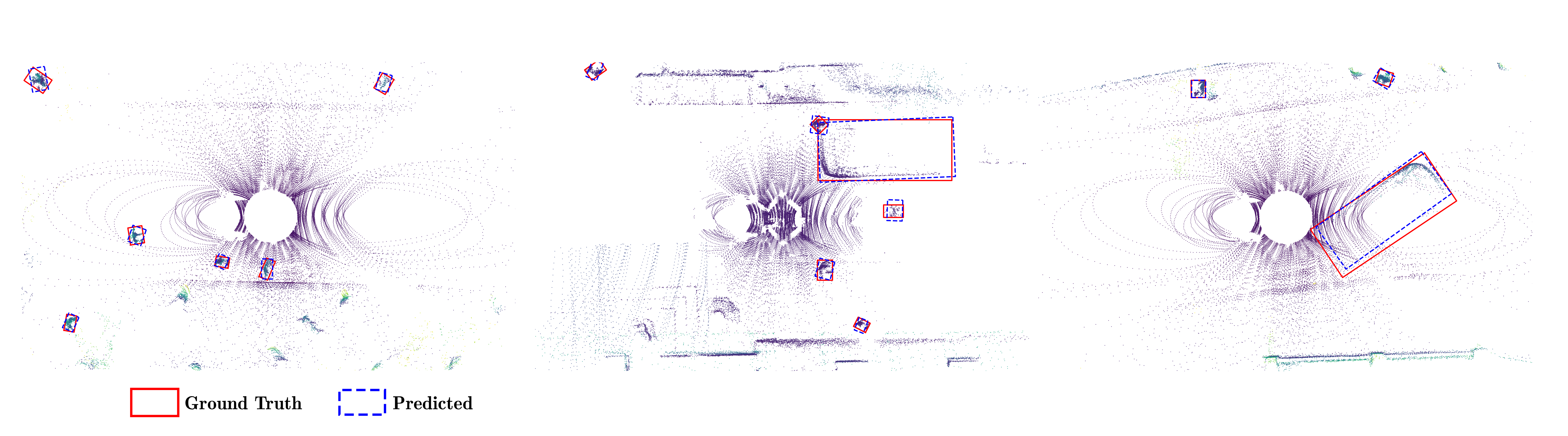}
\caption{
Qualitative BEV visualizations of detection results from the \textit{student-h} model on the Livox-Legged test set. 
Red boxes represent ground-truth annotations, while blue boxes denote predicted 3D bounding boxes
}
\label{fig:livox_vis}
\end{figure*}

\textbf{Evaluation on the Livox-Legged dataset.}
To further assess generalization and embedded deployment capability, we evaluate our models alongside representative baselines on the Livox-Legged dataset.

For a fair comparison, all models on the Livox-Legged dataset are trained from scratch for 36 epochs using the same voxel size, point-cloud range, optimizer, data augmentation strategy, batch size, and training schedule. The training configuration follows the official LION setting. We adopt the fade strategy \mbox{\cite{faded}} to improve performance in the last epoch. All reported results are obtained using the final checkpoint after training.

As shown in Table~\ref{tab:realworld}, our method achieves strong detection performance while demonstrating practical deployment efficiency on an embedded platform, indicating its robustness across different LiDAR configurations and deployment environments. 
\hil{The table also reports the per-class AP on the Livox-Legged test set. Although the \textit{student-h} model achieves a near-saturated Car AP, it also shows substantial improvements on the more challenging Pedestrian and Motorcyclist categories. These results suggest that the proposed knowledge distillation framework effectively preserves detection performance for large, easier objects while enhancing the representation of smaller and more challenging objects, leading to the highest overall mAP.}

Fig.~\ref{fig:livox_vis} presents qualitative BEV visualizations, showing that the proposed \textit{student-h} model produces reasonable 3D bounding box predictions under real-world conditions.
We also observe that smaller objects may exhibit larger orientation errors in the qualitative results. This is mainly because small objects contain fewer LiDAR points and provide weaker geometric cues for heading estimation.

\textbf{Real-world deployment.}

\begin{figure}[b!]
    \centering
    \includegraphics[width=\columnwidth]{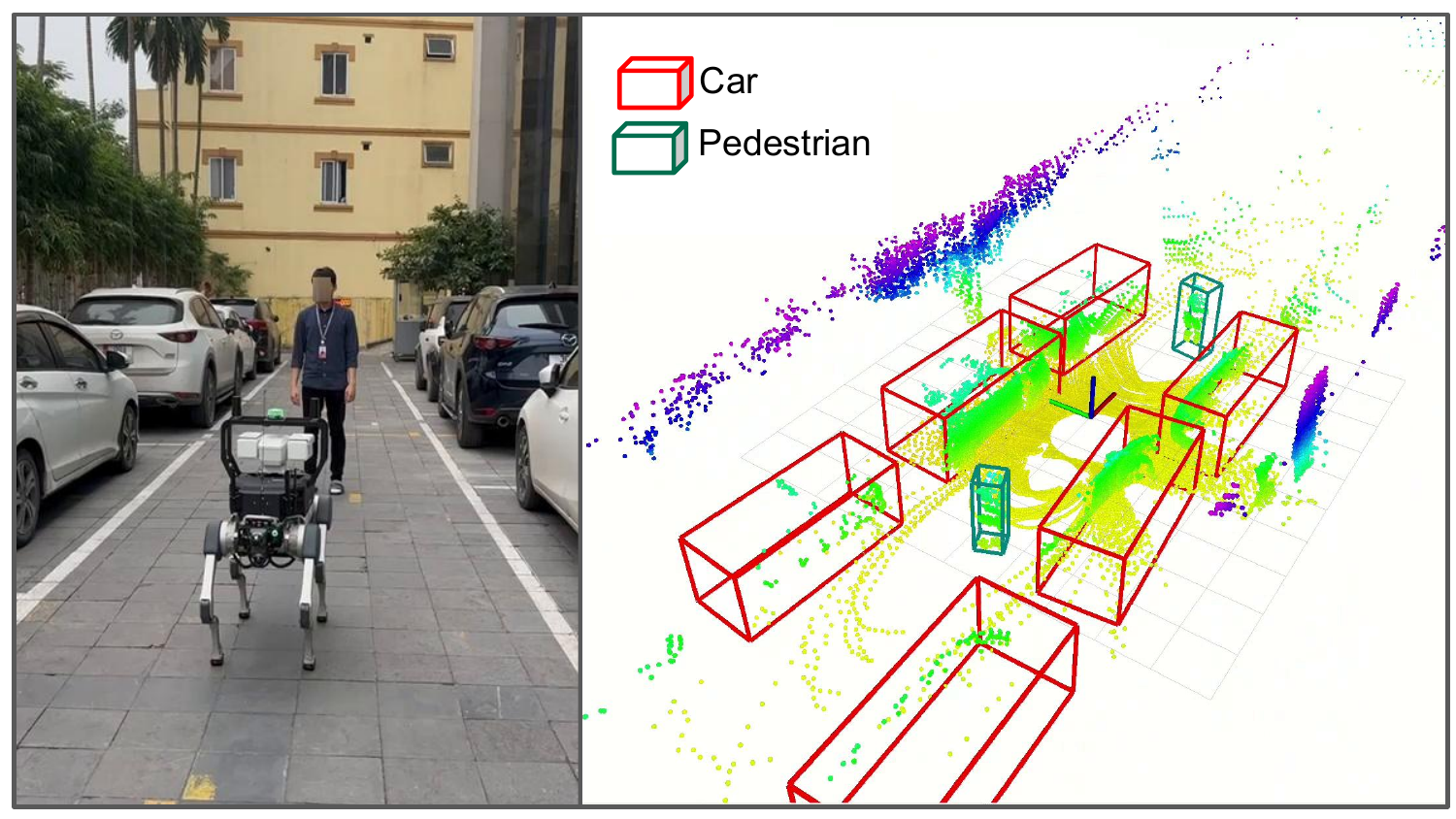}
    \caption{Real-world navigation scenario using the proposed \textit{student-h} model. The pedestrians (in dark teal bounding boxes) and cars (in red bounding boxes) are detected by the 3D perception module. The detection output is integrated into the motion planner, enabling the robot to safely stop when the pedestrian is detected in the forward direction.
    }
    \label{fig:stop}
\end{figure}

To further evaluate onboard deployment, we deploy the trained \textit{student-h} model on a legged patrol robot in a pedestrian-aware navigation scenario. The model runs at approximately 3 Hz on an NVIDIA Jetson Orin NX platform, consistent with the latency reported in Table~{\ref{tab:realworld}}. This inference rate is suitable for low-speed legged patrol scenarios but should not be interpreted as high-frequency real-time perception for fast-moving platforms. The detection output is integrated with the motion planning module, allowing the robot to stop when a pedestrian enters the robot's path. An example of this behavior is shown in Fig.~{\ref{fig:stop}}. The reported latency measures only model inference and does not include the full sensing-to-actuation pipeline, such as sensor acquisition, data transfer, decision making, and low-level control. Further deployment optimization, such as TensorRT acceleration, kernel fusion, and hardware-specific inference optimization, may reduce the end-to-end latency in future work.

\subsection{Ablation studies}

Ablation studies are conducted to understand the effect of different feature-transfer designs on the nuScenes validation set using the \textit{student-d} model trained with 20\% of the training data. For a fair comparison, all projection modules are designed with comparable channel dimensions. The MLP projector consists of three fully connected layers. The Transformer and Mamba projectors employ a lightweight encoder structure composed of an input projection layer, a single Transformer/Mamba block, and an output projection layer. The convolutional projector uses a lightweight $(1\times1)$ convolution network to align the student feature dimension with the teacher feature space. Table~{\ref{tab:feature_ablation}} evaluates the proposed box-aware feature transfer strategy. Without knowledge distillation, the student model achieves 58.54 mAP. Applying feature distillation over all sparse voxel features improves performance to 59.50 mAP, indicating that teacher supervision is beneficial even without object-aware selection. 

\hil{Incorporating the proposed box-aware feature selection further improves performance to 63.00 mAP, demonstrating that selectively transferring object-region features is the primary contributor to the performance gain. Among the evaluated projection modules, the Mamba projector achieves the highest performance, while the MLP, Transformer, and $1\times1$ convolution projectors also provide competitive improvements. These results suggest that feature-space alignment is important for effective knowledge transfer, whereas the choice of projection module provides a relatively modest additional improvement in our setting.}

Table~{\ref{tab:teacher_branch_ablation}} highlights the contribution of different branches to the proposed multi-branch teacher architecture. When trained on the full nuScenes training set and evaluated on the validation set, the shallow-only and deep-only Mamba branches achieve similar performance, with 67.39 and 67.52 mAP, respectively. Combining the two branches yields the best performance, reaching 68.22 mAP and 72.13 NDS. 
\hil{These results suggest that integrating both shallow and deep branches can provide complementary feature representations, leading to a modest but consistent improvement over using either branch alone as the teacher for knowledge distillation.}

\begin{table}[h!]
\centering
\caption{Comparison of Different Feature-transfer Designs and Projection Modules}
\label{tab:feature_ablation}
\resizebox{0.5\textwidth}{!}{
\begin{tabular}{lcccc}
\hline
Method & Box Aware & Projector Type & mAP (\%) $\uparrow$ & NDS (\%) $\uparrow$ \\
\hline
Student w/o KD & - & - & 58.54 & 65.25 \\
Naive feature KD (all voxels)& \xmark & Mamba Encoder & 59.50 & 66.63 \\
Box-aware KD + 1$\times$1 Conv & \checkmark & 1$\times$1 Conv & 61.85 & 67.23 \\
Box-aware KD + MLP& \checkmark & MLP & 62.28 & 68.25\\
Box-aware KD + Transformer & \checkmark & Transformer & 61.40 & 67.65 \\
Ours & \checkmark & Mamba Encoder & 63.00 & 68.59 \\
\hline
\end{tabular}
}
\end{table}

\begin{table}[h!]
\centering
\caption{Effect of Shallow and Deep Mamba Branches in the Teacher Model}
\label{tab:teacher_branch_ablation}
\resizebox{0.4\textwidth}{!}{
\begin{tabular}{lccc}
\hline
Teacher Configuration & mAP (\%) $\uparrow$ &  NDS (\%) $\uparrow$ \\
\hline
Shallow Mamba only & 67.39  & 71.78  \\
Deep Mamba only & 67.52 & 71.96  \\
Shallow + Deep Mamba & 68.22  &  72.13  \\
\hline
\end{tabular}
}
\end{table}
\section{Conclusions} \label{sec:Conclusions}
In this work, we proposed a novel Mamba-based knowledge distillation framework for efficient LiDAR-based 3D object detection in robotic perception systems. A Multi-branch Mamba 3D Backbone is designed as a high-capacity teacher model to enhance multi-scale feature representation, which is followed by a Mamba adaptive distillation strategy that performs object-aware latent state alignment in voxel space for effective knowledge transfer to compact student models. To evaluate student models, we reformulated the CPR metric by incorporating inference latency. Extensive experiments on both the public nuScenes dataset and our real-world dataset demonstrate that the distilled student models achieve competitive detection performance while significantly reducing model size and inference time. In particular, the most efficient student variant preserves the teacher’s accuracy while operating at substantially lower latency, making it suitable for resource-constrained robotic platforms. These results suggest that Mamba-based 3D detectors can be compressed through box-aware voxel-space distillation, enabling efficient onboard 3D perception on resource-constrained robotic platforms. Future work will explore extensions to non-Mamba sparse voxel backbones and deployment-oriented acceleration. A limitation of the proposed box-aware distillation strategy is that it relies on ground-truth bounding boxes during supervised training. This should be addressed in future extensions.

\bibliographystyle{IEEEtran}
\bibliography{main}

@String(CVPR= {IEEE Conf. Comput. Vis. Pattern Recog.})

@String(ICLR = {Int. Conf. Learn. Represent.})

@String(CVPR  = {CVPR})

@String(ICLR  = {ICLR})

@InProceedings{faded,
    author    = {Wang, Chunwei and Ma, Chao and Zhu, Ming and Yang, Xiaokang},
    title     = {PointAugmenting: Cross-Modal Augmentation for 3D Object Detection},
    booktitle = {Proceedings of the IEEE/CVF Conference on Computer Vision and Pattern Recognition (CVPR)},
    month     = {June},
    year      = {2021},
    pages     = {11794-11803}
}

@INPROCEEDINGS{mimic,
  author={Li, Quanquan and Jin, Shengying and Yan, Junjie},
  booktitle={2017 IEEE Conference on Computer Vision and Pattern Recognition (CVPR)}, 
  title={Mimicking Very Efficient Network for Object Detection}, 
  year={2017},
  volume={},
  number={},
  pages={7341-7349},
  doi={10.1109/CVPR.2017.776}}

@inproceedings{
voxelmamba,
title={\hil{Voxel Mamba: Group-Free State Space Models for Point Cloud based 3D Object Detection}},
author={Guowen Zhang and Lue Fan and Chenhang HE and Zhen Lei and Zhaoxiang Zhang and Lei Zhang},
booktitle={The Thirty-eighth Annual Conference on Neural Information Processing Systems},
year={2024}
}

@INPROCEEDINGS{voxelnext,
  author={Chen, Yukang and Liu, Jianhui and Zhang, Xiangyu and Qi, Xiaojuan and Jia, Jiaya},
  booktitle={2023 IEEE/CVF Conference on Computer Vision and Pattern Recognition (CVPR)}, 
  title={\hil{VoxelNeXt: Fully Sparse VoxelNet for 3D Object Detection and Tracking}}, 
  year={2023},
  volume={},
  number={},
  pages={21674-21683},
  doi={10.1109/CVPR52729.2023.02076}}

@INPROCEEDINGS{fg,
  author={Wang, Tao and Yuan, Li and Zhang, Xiaopeng and Feng, Jiashi},
  booktitle={2019 IEEE/CVF Conference on Computer Vision and Pattern Recognition (CVPR)}, 
  title={\hil{Distilling Object Detectors With Fine-Grained Feature Imitation}}, 
  year={2019},
  volume={},
  number={},
  pages={4928-4937},
  doi={10.1109/CVPR.2019.00507}}

@INPROCEEDINGS{transfusionl,
  author={Bai, Xuyang and Hu, Zeyu and Zhu, Xinge and Huang, Qingqiu and Chen, Yilun and Fu, Hangbo and Tai, Chiew-Lan},
  booktitle={2022 IEEE/CVF Conference on Computer Vision and Pattern Recognition (CVPR)}, 
  title={\hil{TransFusion: Robust LiDAR-Camera Fusion for 3D Object Detection with Transformers}}, 
  year={2022},
  volume={},
  number={},
  pages={1080-1089},
  doi={10.1109/CVPR52688.2022.00116}}

@article{liu2024lion,
  title   = {LION: Linear Group RNN for 3D Object Detection in Point Clouds},
  author  = {Zhe Liu and Jinghua Hou and Xingyu Wang and Xiaoqing Ye and Jingdong Wang and Hengshuang Zhao and Xiang Bai},
  journal = {Advances in Neural Information Processing Systems},
  year    = {2024}
}

@inproceedings{spartialkd,
  title     = {Towards Efficient 3D Object Detection with Knowledge Distillation},
  author    = {Yang, Jihan and Shi, Shaoshuai and Ding, Runyu and Wang, Zhe and Qi, Xiaojuan},
  booktitle = {Advances in Neural Information Processing Systems},
  year      = {2022}
}

@inproceedings{pvrcnn,
  author={Shi, Shaoshuai and Guo, Chaoxu and Jiang, Li and Wang, Zhe and Shi, Jianping and Wang, Xiaogang and Li, Hongsheng},
  booktitle={2020 IEEE/CVF Conference on Computer Vision and Pattern Recognition (CVPR)}, 
  title={PV-RCNN: Point-Voxel Feature Set Abstraction for 3D Object Detection}, 
  year={2020},
  volume={},
  number={},
  pages={10526-10535},
  doi={10.1109/CVPR42600.2020.01054}
}

@article{Hinton2015DistillingTK,
  title   = {Distilling the Knowledge in a Neural Network},
  author  = {Geoffrey E. Hinton and Oriol Vinyals and Jeffrey Dean},
  journal = {ArXiv},
  year    = {2015}
}

@misc{openpcdet2020,
  author       = {OpenPCDet Development Team},
  title        = {{OpenPCDet: An Open-Source Toolbox for 3D Object Detection from Point Clouds}},
  year         = {2020},
}

@inproceedings{romero2015fitnetshintsdeepnets,
  author       = {Adriana Romero and
                  Nicolas Ballas and
                  Samira Ebrahimi Kahou and
                  Antoine Chassang and
                  Carlo Gatta and
                  Yoshua Bengio},
  title        = {\hil{FitNets: Hints for Thin Deep Nets}},
  booktitle    = {3rd International Conference on Learning Representations, {ICLR}},
  year         = {2015},
}

@inproceedings{transformer,
 author = {Vaswani, Ashish and Shazeer, Noam and Parmar, Niki and Uszkoreit, Jakob and Jones, Llion and Gomez, Aidan N and Kaiser, \L ukasz and Polosukhin, Illia},
 booktitle = {Advances in Neural Information Processing Systems},
 publisher = {Curran Associates, Inc.},
 title = {\hil{Attention is All You Need}},
 volume = {30},
 year = {2017}
}

@inproceedings{nuscenes,
  author    = {Holger Caesar and Varun Bankiti and Alex H. Lang and Sourabh Vora and Venice Erin Liong and Qiang Xu and Anush Krishnan and Yu Pan and Giancarlo Baldan and Oscar Beijbom},
  title     = {{nuScenes: A Multimodal Dataset for Autonomous Driving}},
  booktitle = {Proceedings of the IEEE/CVF Conference on Computer Vision and Pattern Recognition (CVPR)},
  pages     = {11621--11631},
  year      = {2020}
}

@InProceedings{waymo,
author = {Sun, Pei and Kretzschmar, Henrik and Dotiwalla, Xerxes and Chouard, Aurelien and Patnaik, Vijaysai and Tsui, Paul and Guo, James and Zhou, Yin and Chai, Yuning and Caine, Benjamin and Vasudevan, Vijay and Han, Wei and Ngiam, Jiquan and Zhao, Hang and Timofeev, Aleksei and Ettinger, Scott and Krivokon, Maxim and Gao, Amy and Joshi, Aditya and Zhang, Yu and Shlens, Jonathon and Chen, Zhifeng and Anguelov, Dragomir},
title = {\hil{Scalability in Perception for Autonomous Driving: Waymo Open Dataset}},
booktitle = {Proceedings of the IEEE/CVF Conference on Computer Vision and Pattern Recognition (CVPR)},
month = {June},
year = {2020}
}

@article{yan2018second,
  author    = {Yan Yan and Yuxing Mao and Bo Li},
  title     = {{SECOND: Sparsely Embedded Convolutional Detection}},
  journal   = {Sensors},
  volume    = {18},
  number    = {10},
  pages     = {3337},
  year      = {2018},
  publisher = {MDPI},
  doi       = {10.3390/s18103337}
}

@inproceedings{lang2019pointpillars,
  author    = {Alex H. Lang and Sourabh Vora and Holger Caesar and Lubing Zhou and Jiong Yang and Oscar Beijbom},
  title     = {{PointPillars: Fast Encoders for Object Detection from Point Clouds}},
  booktitle = {Proceedings of the IEEE Conference on Computer Vision and Pattern Recognition (CVPR)},
  pages     = {12697--12705},
  year      = {2019}
}

@inproceedings{qi2017pointnetplusplus,
  author    = {Charles Ruizhongtai Qi and Li Yi and Hao Su and Leonidas J. Guibas},
  title     = {{PointNet++: Deep Hierarchical Feature Learning on Point Sets in a Metric Space}},
  booktitle = {Advances in Neural Information Processing Systems (NeurIPS)},
  pages     = {5099--5108},
  year      = {2017}
}

@inproceedings{yin2021centerbased,
  author    = {Tianwei Yin and Xingyi Zhou and Philipp Kr{\"a}henb{\"u}hl},
  title     = {{Center-Based 3D Object Detection and Tracking}},
  booktitle = {Proceedings of the IEEE/CVF Conference on Computer Vision and Pattern Recognition (CVPR)},
  pages     = {11784--11793},
  year      = {2021}
}

@inproceedings{wang2023dsvt,
  author    = {Haiyang Wang and Chen Shi and Shaoshuai Shi and Meng Lei and Sen Wang and Di He and Bernt Schiele and Liwei Wang},
  title     = {{DSVT: Dynamic Sparse Voxel Transformer with Rotated Sets}},
  booktitle = {Proceedings of the IEEE/CVF Conference on Computer Vision and Pattern Recognition (CVPR)},
  pages     = {13520--13529},
  year      = {2023}
}

@INPROCEEDINGS{zhang2022pointdistillerstructuredknowledgedistillation,
author = { Zhang, Linfeng and Dong, Runpei and Tai, Hung-Shuo and Ma, Kaisheng },
booktitle = { 2023 IEEE/CVF Conference on Computer Vision and Pattern Recognition (CVPR) },
title = {\hil{ PointDistiller: Structured Knowledge Distillation Towards Efficient and Compact 3D Detection }},
year = {2023},
volume = {},
ISSN = {},
pages = {21791-21801},
doi = {10.1109/CVPR52729.2023.02087},
publisher = {IEEE Computer Society},
}

@INPROCEEDINGS{itkd,
  author={Cho, Hyeon and Choi, Junyong and Baek, Geonwoo and Hwang, Wonjun},
  booktitle={2023 IEEE/CVF Conference on Computer Vision and Pattern Recognition (CVPR)}, 
  title={\hil{itKD: Interchange Transfer-based Knowledge Distillation for 3D Object Detection}}, 
  year={2023},
  volume={},
  number={},
  pages={13540-13549},
  doi={10.1109/CVPR52729.2023.01301}}

@inproceedings{
s6,
title={Mamba: Linear-Time Sequence Modeling with Selective State Spaces},
author={Albert Gu and Tri Dao},
booktitle={First Conference on Language Modeling},
year={2024}}

@article{kitti,
  author  = {A Geiger and P Lenz and C Stiller and R Urtasun},
  title   = {Vision meets robotics: The KITTI dataset},
  journal = {The International Journal of Robotics Research},
  volume  = {32},
  number  = {11},
  pages   = {1231-1237},
  year    = {2013},
  doi     = {10.1177/0278364913491297}
}

@inproceedings{LSSL,
  author = {Gu, Albert and Johnson, Isys and Goel, Karan and Saab, Khaled and Dao, Tri and Rudra, Atri and Re, Christopher},
  title = {Combining Recurrent, Convolutional, and Continuous-time Models with Linear State Space Layers},
  booktitle = {Advances in Neural Information Processing Systems (NeurIPS)},
  year = {2021}
}

@inproceedings{S4D,
  author = {Gu, Albert and Goel, Karan and Gupta, Ankit and Re, Christopher},
  title = {On the Parameterization and Initialization of Diagonal State Space Models},
  booktitle = {Advances in Neural Information Processing Systems (NeurIPS)},
  year = {2022}
}

@inproceedings{DSS,
  author = {Gupta, Ankit and Gu, Albert and Berant, Jonathan},
  title = {Diagonal State Spaces are as Effective as Structured State Spaces},
  booktitle = {Advances in Neural Information Processing Systems (NeurIPS)},
  year = {2022}
}

@InProceedings{vim,
  title = 	 {\hil{Vision Mamba: Efficient Visual Representation Learning with Bidirectional State Space Model}},
  author =       {Zhu, Lianghui and Liao, Bencheng and Zhang, Qian and Wang, Xinlong and Liu, Wenyu and Wang, Xinggang},
  booktitle = 	 {Proceedings of the 41st International Conference on Machine Learning},
  pages = 	 {62429--62442},
  year = 	 {2024},
  volume = 	 {235},
  series = 	 {Proceedings of Machine Learning Research},
  month = 	 {21--27 Jul},
  publisher =    {PMLR},
}

@inproceedings{vmamba,
 author = {Liu, Yue and Tian, Yunjie and Zhao, Yuzhong and Yu, Hongtian and Xie, Lingxi and Wang, Yaowei and Ye, Qixiang and Jiao, Jianbin and Liu, Yunfan},
 booktitle = {Advances in Neural Information Processing Systems},
 doi = {10.52202/079017-3273},
 pages = {103031--103063},
 publisher = {Curran Associates, Inc.},
 title = {\hil{VMamba: Visual State Space Model}},
 volume = {37},
 year = {2024}
}

\newpage

\vfill

\end{document}